\documentclass{article}

\usepackage{microtype}
\usepackage{graphicx}
\usepackage{subcaption}
\usepackage{booktabs} 
\usepackage{multirow}
\usepackage[most]{tcolorbox}
\usepackage{enumitem}

\usepackage{hyperref}

\usepackage[accepted]{icml2026}

\usepackage{amsmath}
\usepackage{amssymb}
\usepackage{mathtools}
\usepackage{amsthm}

\usepackage[capitalize,noabbrev]{cleveref}

\theoremstyle{plain}
\newtheorem{theorem}{Theorem}[section]

\theoremstyle{definition}

\theoremstyle{remark}
\newtheorem{remark}[theorem]{Remark}

\DeclareMathOperator*{\E}{{\mathbb{E}}}

\usepackage[textsize=tiny]{todonotes}

\icmltitlerunning{Efficient Reasoning with Hidden Thinking}

\begin{document}

\twocolumn[
  \icmltitle{
  Efficient Reasoning with Hidden Thinking
  }



  \icmlsetsymbol{equal}{*}

  \begin{icmlauthorlist}
    \icmlauthor{Xuan Shen}{zju}
    \icmlauthor{Yizhou Wang}{adobe}
    \icmlauthor{Yufa Zhou}{duke}
    \icmlauthor{Xiangxi Shi}{osu}
    \icmlauthor{Pu Zhao}{neu}
    \icmlauthor{Yanzhi Wang}{neu}
    \icmlauthor{Jiuxiang Gu}{adobe}
  \end{icmlauthorlist}

  \icmlaffiliation{zju}{College of Computer Science and Technology, Zhejiang University, Hangzhou, Zhejiang, China}
  \icmlaffiliation{adobe}{Adobe, San Jose, CA, USA}
  \icmlaffiliation{duke}{Department of Computer Science, Duke University, Durham, NC, USA}
  \icmlaffiliation{osu}{Oregon State University, Corvallis, OR, USA}
  \icmlaffiliation{neu}{Department of Electrical and Computer Engineering, Northeastern University, Boston, MA, USA}

  \icmlcorrespondingauthor{Pu Zhao}{p.zhao@northeastern.edu}
  \icmlcorrespondingauthor{Jiuxiang Gu}{gu.jiuxiang@gmail.com}

  \icmlkeywords{Machine Learning, ICML}

  \vskip 0.3in
]



\printAffiliationsAndNotice{}  

\begin{abstract}

Chain-of-Thought (CoT) reasoning has become a powerful framework for improving complex problem-solving capabilities in Multimodal Large Language Models (MLLMs).
However, the verbose nature of textual reasoning introduces significant inefficiencies.
In this work, we propose \textbf{Heima} (as hidden llama), an effective CoT compression framework that condenses lengthy CoTs into a small set of abstract thinking tokens, preserving essential reasoning while removing redundancy.
We then conduct a theoretical analysis from an information-theoretic perspective, quantifying the information gap induced by compression, showing that reasoning capability is preserved when non-trivial mutual information is retained.
To further explore and quantify this information gap, we design the adaptive interpreter that maps thinking tokens back to variable-length textual sequences, thereby reconstructing the reasoning process.
Experiments across diverse reasoning benchmarks demonstrate that Heima improves reasoning efficiency, while maintaining or even achieving better zero-shot accuracy.
Moreover, the interpreter reconstructs coherent reasoning progresses from compressed thinking tokens, revealing that the information gap is minimal and validating the effectiveness of the proposed framework.
This work paves the way for scalable latent reasoning models and advances our understanding of efficient reasoning processes in large models.
Code: \url{https://github.com/shawnricecake/Heima}

\end{abstract}

\section{Introduction}

The recent rise in popularity of Multimodal Large Language Models (MLLMs)~\citep{achiam2023gpt4_tech_report, qwen-VL, liu2024llava, lai2024lisa, xu2024llavacot, shen2023deepmad,shen2025quart,shen2025sparse,shen2026oida}, which integrate vision techniques with traditional Large Language Models (LLMs), has spurred interest in leveraging Chain-of-Thought (CoT)~\citep{wei2022original_cot} reasoning to enhance their capabilities for solving complex problems.
It not only enhances interpretability but also enables more effective multi-step reasoning, equipping MLLMs to address tasks that demand intricate logical understanding and contextual coherence, especially when processing the inherent complexity of visual information.
However, reasoning with CoT often requires generating a substantial amount of additional reasoning texts, particularly for complex problems, leading to expensive inference costs.
Thus, efficiency has become a central theme in applying MLLMs, making it crucial to reduce the number of tokens generated during reasoning to enhance overall computational performance.

Recent works~\citep{hao2024meta_compress_cot, deng2024compress_cot} explore the latent reasoning methods.
Approach~\citep{hao2024meta_compress_cot} explores compressing CoTs for a small-scale model, GPT-2~\citep{radford2019gpt2}, on individual reasoning tasks in the text-only setting. However, this leaves a significant gap in extending CoT compression to large-scale MLLMs that must handle general reasoning tasks with multimodal inputs. 
Other works~\citep{pi2023detgpt, yan2024visa, deng2024motion} investigate latent reasoning in MLLMs by employing visual decoders for segmentation, detection, and recognition, aiming to decode latent information in MLLM generated tokens.
This shows the feasibility of latent-space reasoning and motivates deeper investigation into its capabilities.

\begin{figure*}[t]
  \centering
  \includegraphics[width=0.99\linewidth]{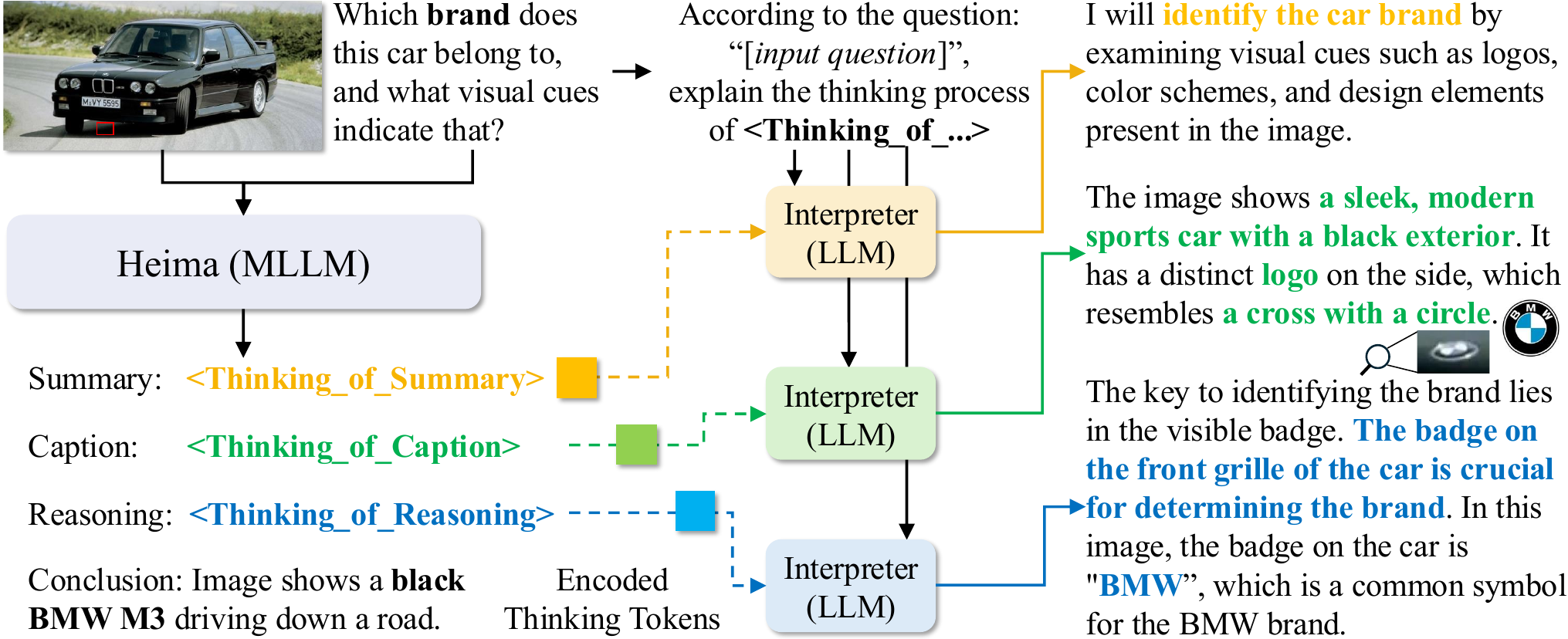}
  \caption{
  Heima (MLLM) peforms reasoning with thinking tokens based on image and question.
  Compressed thinking tokens and question are then fed to interpreters (LLMs) for CoT reconstruction.
  In reconstructed reasoning progress, the caption interpreter successfully retrieves image information, describing it as \textit{"The image shows a sleek, modern sports car with a black exterior"} and identifying a distinct feature of logo as \textit{"a cross with a circle"}. 
  Also, the reasoning interpreter accurately deduces the logo as BMW based on this distinctive symbol.
  \textbf{This verifies that interpreters effectively reconstruct visual features from pure textual inputs.}
  }
  \label{fig:introduction_whole_pipeline}
  \vspace{-4mm}
\end{figure*}

In this work, we propose Heima, the first CoT compression framework for MLLMs to achieve efficient reasoning.
Instead of relying on verbose CoTs, Heima performs reasoning in the latent space, compressing CoTs into compact thinking tokens.
First, we design  Heima as the CoT compressor of our framework.
Specifically, we train the reasoning MLLM to distill each CoT stage into a single thinking token, \texttt{<CoT>}, utilizing step-by-step distillation for more effective mapping.
Then, we provide theoretical analysis from the information-theoretic perspective to quantify the information gap between textual CoTs and thinking tokens, showing that reasoning effectiveness is preserved when non-trivial mutual information is retained.
To further assess and quantify this gap, we develop corresponding interpreters fine-tuned from LLMs that adaptively decode thinking tokens into textual sequences of varying length.
By leveraging explanatory prompts, the interpreters effectively reconstruct reasoning progress from compressed representations.

The whole framework is shown in Figure~\ref{fig:introduction_whole_pipeline}.
We accelerate the reasoning using Heima, which performs reasoning in latent space by generating a significantly reduced number of thinking tokens instead of verbose textual CoTs, thereby producing answers more efficiently.
The last hidden states of these thinking tokens are further delivered to interpreters for the reconstruction of the textual reasoning sentences.
Experimental results demonstrate that our approach significantly enhances reasoning efficiency by generating far fewer tokens while achieving comparable or even superior performance on a list of zero-shot reasoning benchmarks.
Furthermore, the results demonstrate that the reasoning progress reconstructed by interpreters, even without visual information, closely align with the original CoTs texts generated based on multimodal inputs.
These findings establish that the information gap induced by compression is negligible and affirm the effectiveness as well as the interpretability of the thinking tokens produced by Heima.
Our contributions are summarized below:
\begin{itemize}
\vspace{-4mm}

\item \textbf{1.} We propose Heima, the first reasoning acceleration framework for MLLMs that conducts reasoning in latent space by generating compact thinking tokens rather than verbose textual CoTs.

\vspace{-2mm}
\item \textbf{2.} We develop an information-theoretic analysis of CoT compression within Heima that quantifies the compression-induced information gap and proves that reasoning capability is preserved under retention of non-trivial mutual information.

\vspace{-2mm}
\item \textbf{3.} We design interpreters with pure LLMs to reconstruct textual reasoning with thinking tokens, which serves to analyze the information gap induced by compression relative to the original CoTs.

\vspace{-2mm}
\item \textbf{4.} Experiments show that Heima attains substantial improvements in reasoning efficiency without sacrificing performance relative to textual CoTs. In addition, reconstructions with interpreters indicate that the information gap induced by compression is negligible, providing empirical validation of the theoretical guarantees underlying the Heima framework.


\end{itemize}

\vspace{-3mm}
\section{Related Work}

\subsection{Chain-of-Thought Reasoning}
With the theoretically validated effectiveness in recent works~\citep{merrill2023expresssive, feng2024towards,zhou2025geometry,wang2025cautious}, CoTs have gained increasing popularity and are  widely adopted as an enhancement method for generating intermediate reasoning processes before arriving at the final answer.
The works~\citep{wei2022original_cot, khot2022decomposed, zhou2022least} focus on designing the effective prompts which decompose the question into a group of reasoning steps for LLMs. The  works~\citep{yue2023mammoth, yu2023metamath, wang2023math, shao2024deepseekmath, liu2024deepseekv3} adopt additional fine-tuning to guide the model to generate reasoning chains.
Meanwhile, to further enhance reasoning performance and make model reasoning more human-like, recent works~\citep{xie2024self, gandhi2024stream, su2024dualformer} have integrated additional search algorithms to improve the relevance of CoT generation and the input content.
However, almost all of these works enhance reasoning performance by generating additional textual tokens, which incurs significant computational costs for large generative models with billions of parameters. 
This motivates us to explore  token reduction methods during  reasoning process.

\subsection{Textual Efficient Reasoning}

Recent works~\citep{ning2023skeleton, kou2024cllms, zhang2023fast, zhan2024exploring,kong2025,li2024eagle,shen2024agile,shen2024hota,shen2024search,shen2025numerical,zhao2025open,lin2025vote,shen2025lazydit,shen2026fastcar,shen2025iccad} aim to accelerate the reasoning process by employing parallel generation through templates or Jacobi decoding, which introduces additional overhead during model inference. 
Meanwhile, the works~\citep{ge2024incontext, chevalier2023adapting, qin2023nugget, liu2023cachegen, munkhdalai2024leave, zhan-etal-2024-rethinking-token,add-Liu-2025-toward, add-liu-rora, add-liu-tsla, add-nguyen-etal-2025-survey, add-zhao-etal-2024-pruning}  adopt contextual compression methods to achieve the efficient generation of the next following contexts based on the previous contexts.
On the other hand, the work~\citep{cheng2024compressed} compresses the CoT into a short sequence of continuous embeddings for the acceleration of reasoning process. 
However, its results on the math dataset reveal a significant degradation in accuracy, indicating the limitations and ineffectiveness of this approach.
Thus, there is still a gap in developing efficient reasoning techniques for large models that  maintain the  advantages of reasoning with CoTs, which motivates us to explore better compression methods with CoTs for higher accuracy and efficiency.

\subsection{Reasoning in Latent Space}

Works~\citep{hao2024meta_compress_cot, yang2024large, biran2024hopping, cheng2024compressed} adopt latent reasoning for LLMs.  
For example, the work~\citep{hao2024meta_compress_cot} addresses the compression of small models (GPT-2) on math datasets with CoTs.  However, its effectiveness remains unverified as the evaluation is limited to math datasets and small model size.
Some works~\citep{liu2024llava, lai2024lisa} include visual information into latent space for MLLMs to enhance the textual reasoning.
The work~\citep{lai2024lisa} employs fine-tuning for both LLMs and segmentation decoders, demonstrating the potential to decode visual information in tokens generated by LLMs. 
Subsequent works~\citep{pi2023detgpt, deng2024motion, yan2024visa} continue to investigate using MLLMs to generate tokens for visual downstream tasks, rather than exploring the construction of internal feature representations within MLLMs for higher-level feature embedding.
The absence of feature construction in MLLMs motivates us to explore the development of latent representations for these models.

\section{Methodology}

We mainly present Heima framework in this section.
First, we outline the training setup of Heima, covering dataset construction and the distillation strategy.
Then, we provide theoretical analysis of the information gap between textual CoTs and compressed thinking tokens from an information-theoretic perspective.
Next, we introduce the interpreter, which maps thinking tokens back to CoT-like trajectories, to further explore the information gap introduced by compression.
Finally, we demonstrate how this approach enables efficient compressed reasoning at inference time.

\subsection{Heima Framework} \label{sec:encoder}

We introduce Heima to compress verbose textual CoTs into compact thinking tokens through further distillation, enabling faster on-the-fly CoT reasoning.

\paragraph{Dataset Preparation.}

Given $N$ total samples, original CoT training dataset $D$ is defined as follows,
\begin{equation}
D := \big\{ \bigl(X,\, \mathrm{CoTs},\, Y \bigr) \big\}, ~~ |D| = N ,
\end{equation} 
where  
$X$ denotes the visual and query inputs,  
$\mathrm{CoTs}:= \{\mathrm{CoT}_{(k)}\}_{k=1}^{K_i}$ denotes the sequence of $K_i$ textual CoT stages in $i$-th sample, and  
$Y$ denotes the final answers.  




\paragraph{Thinking Token Dataset.}

We define the reasoning MLLM as $\mathcal{H}$, which is fine-tuned on the updated dataset to incorporate CoTs. Specifically,  
we  update the training set $D$ by replacing each $\mathrm{CoTs}$ with thinking tokens---denoted as
$\texttt{<CoTs>}:= \{\texttt{<CoT>}_{(k)}\}_{k=1}^{K_i}$. 
For each thinking token $\texttt{<CoT>}_{(k)}$ at the $k$-th stage, it is defined with \textit{a unique special token and added to the vocabulary} to enable explicit textual visualization of the reasoning process. For example, in Figure~\ref{fig:introduction_whole_pipeline}, we define a new token $\texttt{<Thinking\_{of}\_Summary>}$ in vocabulary as the thinking token for the summary stage.
Note that different samples with varying values of $i$ share the same thinking token $\texttt{<CoT>}_{(k)}$ at the same stage $k$. 
The updated Heima dataset $D_H$ is then defined as follows,
\begin{equation}
D_H := \bigl\{
   \bigl( X,  
   \texttt{<CoTs>},\, 
   Y\bigr)
\bigr\}, ~~ |D_H| = N.
\end{equation}

\paragraph{Distillation Objective.}
To perform the reasoning with thinking tokens in latent space, we further perform fine-tuning using CoT distillation.
The distillation objective is defined as follows,
\begin{equation}\label{eq:fine-tune}
\mathcal{L}(\theta)
= - \E_{(X, Y, \texttt{<CoTs>}) \sim D_H} \log P_\theta(\texttt{<CoTs>}, Y \mid X).
\end{equation} 
Through this distillation, we fine-tune model to directly predict thinking tokens $\texttt{<CoTs>}$ instead of verbose $\mathrm{CoTs}$.


\begin{figure*}[t]
  \centering
  \includegraphics[width=0.9\linewidth]{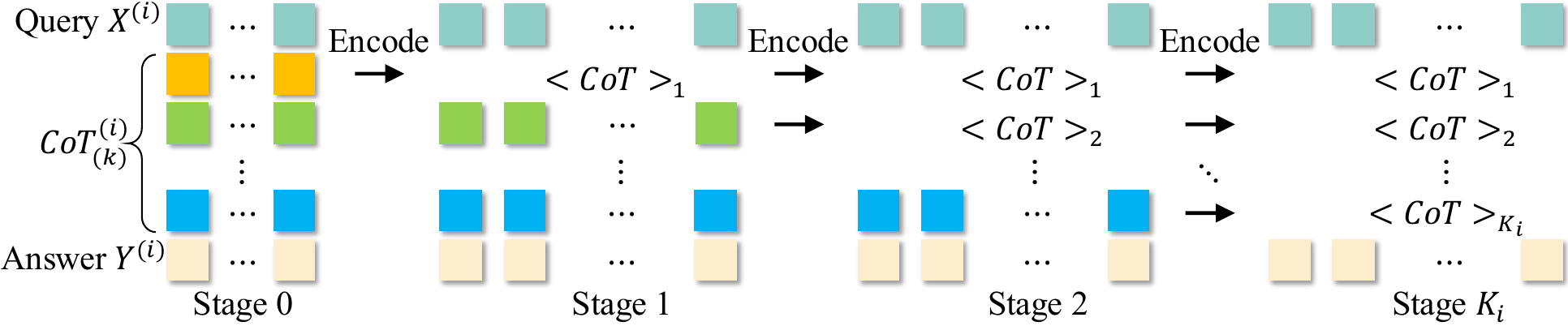}
  \vspace{-1mm}
  \caption{
  Visualization of progressive distillation. Each row shows the training data in that stage.
  }
  \label{fig:method_progressive_encoding}
  \vspace{-3mm}
\end{figure*}

\paragraph{Progressive Distillation.}

To facilitate a seamless transition from textual reasoning to hidden thinking while maintaining model performance, we further adopt the progressive distillation strategy as shown in Figure~\ref{fig:method_progressive_encoding}. 
Specifically, we do not distill all CoT stages into thinking tokens at once. Instead, we \textit{progressively} distill each stage into a thinking token, one by one.
Formally, our progressive distillation has $M := \max\{K_i\}_{i=1}^{N} + 1$ stages. After we finish the $s$-th stage distillation, we move on to its next $(s+1)$-th stage distillation, until reaching the final stage. In the $s$-th stage ($s \in \mathbb{N}, 0 \leq s < M$), we prepare the pregressive training data $|D_P| \le N M$ using 
\begin{equation}
D_P := \bigl\{ 
   \bigl( X,\,  
   \{\texttt{<CoT>}_{(k)}\}_{k=1}^{s} ,\,
   \{\mathrm{CoT}_{(k)}\}_{k=s+1}^{K_i} , \,
   Y \bigr) 
\bigr\}. 
\end{equation} 
For $s=0$, the set $\{\texttt{<CoT>}_{(k)}\}_{k=1}^{s}$ is empty with no thinking tokens in training. 
For $s>0$, we finetune Heima $\mathcal{H}_{\theta}(\cdot)$ as follows, 
\begin{align}\label{eq:stage_mid}
\max_{\theta}
\E 
\log P_{\theta} \Bigl(
    \{\texttt{<CoT>}_{(k)}\}_{k=1}^{s}, \,
      \{\mathrm{CoT}_{(k)}\}_{k=s+1}^{K_i}, \,
      Y 
    \;\big|\; X
\Bigr).
\end{align}
In the above expression, during the $s$-th stage distillation, the first $s$ CoT stages  are distilled with thinking tokens, while the rest CoT stages are untouched and trained with textual reasoning tokens.
As the value of $s$ increases, more CoT stages are distilled into thinking tokens as shown in Figure~\ref{fig:method_progressive_encoding}. 

This approach allows the model to gradually internalize the reasoning processes with multiple CoTs and integrate them into thinking tokens.
After the final progressive stage, we introduce an additional \textit{recovering stage}, where the model is further optimized using only thinking tokens as in Equation~\eqref{eq:fine-tune}. 
This extra recovering stage optimizes the transitions and interactions between the thinking tokens across different stages, ensuring a cohesive alignment of information learned throughout the distillation process.
Additionally, it consolidates the overall learning process, enhancing the model's ability to effectively utilize the distilled reasoning patterns for improved performance and robustness in downstream reasoning tasks.

\paragraph{Efficient Reasoning.}
To accelerate reasoning with Heima framework, we deploy Heima solely as the service reasoning MLLM, benefiting from lower memory usage and faster generation. 
Heima generates the response corresponding to the given input as follows,
\begin{align}
& P_{\theta}
\bigl(
\texttt{<CoTs>}
, \,
Y
\;\big|\; X
\bigr) 
= 
\prod_{t=1}^{T+K_i} P_{\theta}\bigl(w_{t} \;\Big|\; w_{<t}, X\bigr), 
\end{align} 
where the $(w_{1}, w_{2}, \dots, w_{K_i}) = \texttt{<CoTs>}$ denotes the generated thinking tokens, and the $Y = (w_{K_i+1}, w_{K_i+2}, \dots, w_{K_i+T})$ denotes the tokens corresponding to the final answer. 
We obtain the final answer by Heima using only $K_i$ intermediate tokens, significantly reduced from the original $\sum_{k=1}^{K_i} | \mathrm{CoT}_{(k)}|$ tokens, enabling faster generation and avoiding complex \& verbose textual CoTs.

\subsection{Information-Theoretic Compression}\label{sec:theory_analysis}

Distillation (Equation~\eqref{eq:fine-tune}) represents a form of compression~\citep{deletang2024language, huang2024compression}, since such loss induces a code when paired with an ideal entropy coder~\citep{mackay2003information}.
Extending this view, we provide an analysis of information variance across the pre- and post-distillation stages.


\paragraph{Notations.}
For a random variable $X\sim P_X$, entropy is $H(X)=-\mathbb{E}_{x\sim P_X}[\log P_X(x)]$. 
For joint variables $(X,Y)$ with distribution $P_{X,Y}$, conditional entropy is $H(Y\mid X)=\mathbb{E}_{x\sim P_X}[H(P_{Y\mid X=x})]$. 
Mutual information is $I(X;Y)=H(X)-H(X\mid Y)=\mathrm{KL}(P_{X,Y}\,\|\,P_XP_Y)$, 
and conditional mutual information is $I(X;Y\mid Z)=H(X\mid Z)-H(X\mid Y,Z)$.

Given input $X$ for inference, Heima compresses textual $\mathrm{CoTs}$ into compact representations $\texttt{<CoTs>}=f(X,\mathrm{CoTs})$ as thinking tokens with $|\texttt{<CoTs>}|\ll|\mathrm{CoTs}|$. 
The central question is whether $\texttt{<CoTs>}$ retains sufficient task-relevant information about the output $Y$. 
By the data-processing inequality~\citep{mackay2003information} as follows,
\[
0 < I(Y;\texttt{<CoTs>}\mid X) \;\le\; I(Y;\mathrm{CoTs}\mid X),
\]
which shows that $\texttt{<CoTs>}$ carry no more information than $\mathrm{CoTs}$, while still preserving non-trivial information for reasoning.
This motivates Heima as a compressor of CoTs, a property we formalize in Theorem~\ref{thm:heima-info}.

\begin{theorem}[Information preserved under CoT compression]\label{thm:heima-info}
Let $X$ be the input, $\mathrm{CoTs}$ chain-of-thoughts, $\texttt{<CoTs>}=f(X,\mathrm{CoTs})$ thinking tokens, and $Y$ the task output. Since $\texttt{<CoTs>}=f(X,\mathrm{CoTs})$, we have the Markov chain $Y-(X,\mathrm{CoTs})-\texttt{<CoTs>}$. Then
\[
H(Y\mid X,\mathrm{CoTs})\ \le\ H(Y\mid X,\texttt{<CoTs>})\ \le\ H(Y\mid X),
\]
equivalently,
\begin{align*}
\quad 0\ \le\ & I(Y;\texttt{<CoTs>}\mid X)\ \le\ I(Y;\mathrm{CoTs}\mid X),    \\
\text{and} \qquad & I(Y;\mathrm{CoTs}\mid X)-I(Y;\texttt{<CoTs>}\mid X)\\
 &=I(Y;\mathrm{CoTs}\mid X,\texttt{<CoTs>})\ \ge\ 0. 
\end{align*}
Thus, compression into $\texttt{<CoTs>}$ cannot increase the information about $Y$ relative to $\mathrm{CoTs}$. It preserves strictly positive task-relevant information if $I(Y;\texttt{<CoTs>}\mid X) > 0$, and loses no information compared to $\mathrm{CoTs}$ if and only if $I(Y;\mathrm{CoTs}\mid X,\texttt{<CoTs>})=0$ (i.e., $\texttt{<CoTs>}$ is sufficient for $\mathrm{CoTs}$ with respect to $Y$).
\end{theorem}

\begin{proof}
Because $\texttt{<CoTs>}=f(X,\mathrm{CoTs})$, $Y-(X,\mathrm{CoTs})-\texttt{<CoTs>}$ holds, and by the data processing inequality~\citep{mackay2003information} $I(Y;\texttt{<CoTs>}\mid X)\le I(Y;\mathrm{CoTs}\mid X)$. Using the identity $I(Y;W\mid X)=H(Y\mid X)-H(Y\mid X,W)$ for $W\in\{\texttt{<CoTs>},\mathrm{CoTs}\}$ gives $H(Y\mid X,\mathrm{CoTs})\le H(Y\mid X,\texttt{<CoTs>})\le H(Y\mid X)$. Finally, the chain rule with $I(Y;\texttt{<CoTs>}\mid X,\mathrm{CoTs})=0$ yields $I(Y;\mathrm{CoTs}\mid X)=I(Y;\texttt{<CoTs>}\mid X)+I(Y;\mathrm{CoTs}\mid X,\texttt{<CoTs>})$, proving the nonnegative gap and sufficiency condition.
\end{proof}

\begin{remark}\label{rem:info}
This result formalizes the role of CoT compression: the compressed $\texttt{<CoTs>}$ can never be more informative about the target $Y$ than the full chain $\mathrm{CoTs}$, but as long as $I(Y;\texttt{<CoTs>}\mid X) > 0$, it still retains strictly useful information beyond $X$ alone. 
The gap $I(Y;\mathrm{CoTs}\mid X,\texttt{<CoTs>})$ quantifies the reasoning details lost in compression, while the inequality ensures that compressed reasoning remains effective as long as $\texttt{<CoTs>}$ captures nontrivial task-relevant information. 

\end{remark}

Theorem~\ref{thm:heima-info} and Remark~\ref{rem:info} indicate that the gap $I(Y;\mathrm{CoTs}\mid X,\texttt{<CoTs>})$ determines the amount of information loss during the compression from verbose textual CoTs to compact thinking tokens. 
This observation highlights that the effectiveness of reasoning in latent space fundamentally depends on how much of the task-relevant information is preserved after compression.


\subsection{Interpreter Design}\label{sec:decoder}

As explained in the above section, it is essential to analyze and quantify this information gap in order to assess whether the compressed representations remain sufficient for reasoning.
To this end, we further design the corresponding interpreters that reconstruct thinking tokens back into textual reasoning traces, thereby providing an empirical proxy to evaluate the magnitude of this gap and verify the practical validity of reasoning in latent space.
Meanwhile, this can verify if the model is genuinely learning reasoning in latent space rather than merely fitting the data, showing the effectiveness of compressed representations encapsulated within thinking tokens. 
Specifically, for the design of interpreter, we adopt the standard next-token prediction objective in pure LLMs for the reconstruction of variable-length (i.e., adaptive) textual sequences based on thinking tokens.

\paragraph{Adaptive Interpretation.}
After full distillation in Heima, all CoT stages are distilled into thinking tokens. Thinking token for each reasoning stage requires one corresponding interpreter. Thus, to interpret all of the thinking tokens, we train the corresponding interpreters separately.
In detail, we adopt a pretrained LLM as the initialization of interpreter $\mathcal{I}_{\theta_k}(\cdot)$ for the $k$-th CoT stage corresponding to the $k$-th thinking token $\texttt{<CoT>}_{(k)}$, 
where $\theta_k$ denotes its parameters. We do not consider the case of $k=0$ without thinking tokens.
Its training set $D_I$ of the $k$-th stage is designed as follows, for $|D_I| = N,$ let
\begin{align}
D_I := \bigl\{
(
X_{e}, \,
X_q, \,
\texttt{<CoT>}_{(k)}, \,
H_{\texttt{<CoT>}_{(k)}}, \,
\mathrm{CoT}_{(k)}
)\bigr\} , 
\end{align}
where $X_{e}$ denotes the explanatory prompts to guide the model for the interpretation of thinking tokens.
$X_{q}$ denotes the pure textual question. 
$H_{\texttt{<CoT>}_{(k)}}$ denotes the hidden representation (last hidden states) of the thinking token $\texttt{<CoT>}_{(k)}$ generated by Heima.
$\mathrm{CoT}_{(k)}$ denotes the original textual CoT at $k$-th stage. 
Note that, here we only use pure-text LLMs as the interpreters, meaning they cannot read images. The dataset $D_I$ for training interpreters only has text questions without visual images inputs.  

\begin{figure*}[]
  \centering
  \includegraphics[width=0.9\linewidth]{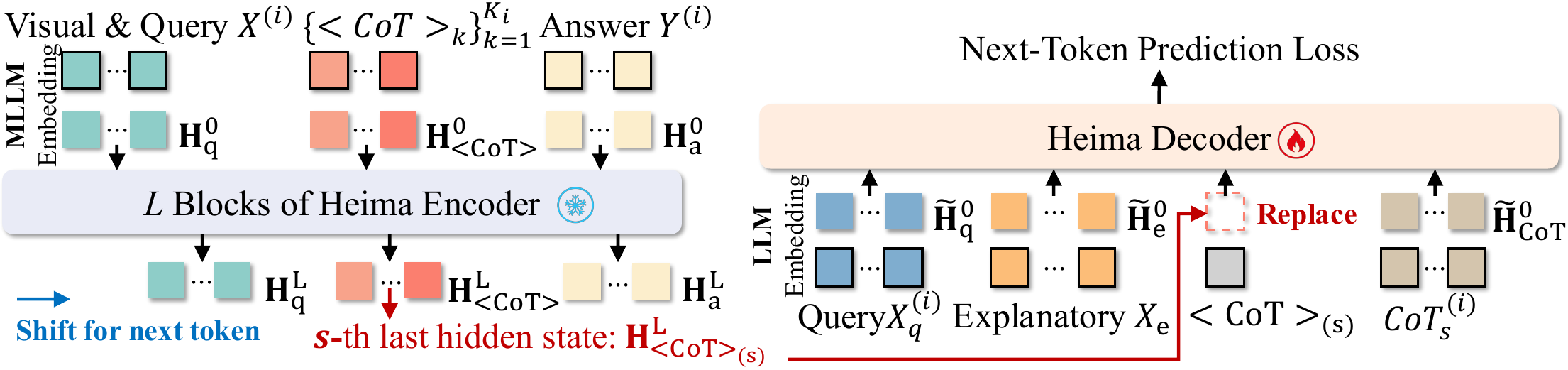}
  \vspace{-1mm}
  \caption{
  Training progress for the interpreter.
  Thinking token is caught from the $s$-th last hidden state of Heima and replaces the embedding of special token $\texttt{<CoT>}_{(s)}$.
  }
  \label{fig:method_adaptive_decoding_train}
  \vspace{-3mm}
\end{figure*}

During the training of interpreter, the frozen Heima is used to generate the thinking tokens. Importantly, we do not feed the token symbol $\texttt{<CoT>}_{(k)}$ directly as input. Instead, we replace it with the corresponding last hidden state $H_{\texttt{<CoT>}_{(k)}}$ from Heima, since the reasoning information is encapsulated in the hidden representation (i.e., last hidden state) rather than the textual symbol. This substitution occurs after the word embedding stage, as illustrated in Figure~\ref{fig:method_adaptive_decoding_train}.
Interpreter is fine-tuned with the next-token prediction loss as follows,
\begin{equation}
\max_{\theta_k}
\E 
\log P_{\theta_k} \Bigl(
    \mathrm{CoT}_{(k)}
    \;
    \big|
    \; 
    X_{e}, \, 
    X_{q}, \,
    H_{\texttt{<CoT>}_{(k)}}
\Bigr).
\end{equation}
%

\paragraph{Explanatory Prompts.}

The single hidden representation $H_{\texttt{<CoT>}_{(k)}}$ alone is insufficient to guide the interpreter $\mathcal{I}_{\theta_k}(\cdot)$ toward reconstructing the original reasoning texts, as language models generally rely on textual instructions to scaffold the generation process. 
Thus, we provide the explanatory prompt for interpreter to enhance usability. 
We use the prompt as follows, 
\begin{quote}
\emph{``According to question: $X_q$, can you explain the thinking progress $\texttt{<CoT>}_{(k)}$?''}
\end{quote}
This kind of prompt ensures that the output reasoning process remains (i) aligned with the original query $X_q$ and (ii) consistent with the hidden representations contained in thinking tokens.

The interpreter is employed to investigate the information gap highlighted in Theorem~\ref{thm:heima-info} and Remark~\ref{rem:info} by comparing the reconstructed reasoning sentences with the original textual CoTs. When the reconstructed reasoning closely aligns in semantics with the original CoTs, the compression-induced information gap is regarded as minimal, thereby confirming that reasoning with thinking tokens preserves the essential reasoning capability.


\begin{table*}[t]
\centering
\caption{
Main results compared to Llama3.2-11B-Vision-Instruct and LLaVA-CoT with both accuracy and number of generated tokens on 6 different multimodal reasoning benchmarks.
}
\vspace{-2mm}
\resizebox{0.99\linewidth}{!}{
\begin{tabular}{l|cccccc|c}
\toprule
Dataset                                                          & MMSar           & MMBench         & MMVet           & MathVista       & AI2D            & Hallusion       & Avg. \\
\midrule
\multirow{2}{*}{Model} & Acc.  & Acc.  & Acc.  & Acc.  & Acc.  & Acc.  & \multirow{2}{*}{Acc.} \\
                     & (\# Token) & (\# Token) & (\# Token) & (\# Token) & (\# Token) & (\# Token) &  \\
\midrule
\begin{tabular}[c]{@{}l@{}}Llama3.2\\ -11B Vision \end{tabular} & 48.1 (140.0)    & 58.2 \ (64.7)     & 50.2 \ (106.0)    & 50.3 \ (240.1)    & 68.5 \ (74.9)     & 37.2 \ (91.4)     & 52.1 \\
\midrule
LLaVA-CoT                                                        & 54.0 \ (181.0)    & 70.7 \ (154.8)    & 49.8 \ (227.2)    & 50.9 \ (216.3)    & 77.6 \ (178.5)    & 63.8 \ (177.9)    & 61.1 \\
\midrule
\begin{tabular}[c]{@{}l@{}}Heima w/o \\ progressive\end{tabular} & 49.7 \ (13.1)     & 72.5 \ (13.3)     & 39.0 \ (71.7)     & 39.3 \ (13.6)     & 75.9 \ (12.6)     & 61.3 \ (15.6)     & 56.3 \\
\midrule
\begin{tabular}[c]{@{}l@{}}Heima w/o \\ recover\end{tabular}     & 49.8 \ (13.0)     & 71.6 \ (13.2)     & 42.8 \ (79.6)     & 39.8 \ (14.0)     & 77.3 \ (12.7)    & 58.5 \ (17.5)     & 56.6 \\
\midrule
\textbf{Heima}                                                   & 49.9 \ (12.8) & 72.8 \ (12.9) & 43.3 \ (75.8) & 43.6 \ (13.8) & 77.5 \ (12.7) & 60.6 \ (16.9) & \textbf{58.0} \\
\bottomrule
\end{tabular}
}
\label{tab:results_main}
\end{table*}

\section{Experimental Results}\label{sec:expeirment}

\subsection{Experiment Setup}

\textbf{Dataset. }
We utilize the \texttt{LLaVA-CoT-100k}~\citep{xu2024llavacot} dataset, a reasoning dataset for MLLMs that integrates samples from several widely used VQA datasets. 
It comprises 100k image-QA pairs with three stages of CoT reasoning: summary, caption, and reasoning.

\textbf{Model Training. }
We adopt the LLaVA-CoT~\citep{xu2024llavacot} pretrained model based on Llama-3.2-11B-Vision-Instruct~\citep{llama32vision} as the initialization of Heima.
The Llama-3.1-8B-Instruct~\citep{llama31} is employed as the initialization of interpreter.
We use torchtune~\citep{torchtune} as the model training framework with LoRA~\citep{hu2021lora} for both Heima and the corresponding interpreter.
During the progressive distillation, we freeze the image encoder component and fine-tune both the decoder module and fusion components of the LLaVA-CoT model. 
This distillation includes the entire attention and MLP modules across all layers, as well as the output projection layer, using a rank of 16, and an alpha of 32.
For the training of interpreter, we apply the same LoRA setting.
Detailed hyperparameters are included in Appendix~\ref{sec:supp_training_hyperparameters}.
The training is conducted on 8$\times$ H100 GPUs.
Besides, to further verify our generalization for different model architectures, the LLaVA-Next-Vicuna-7B~\citep{liu2024llavanext} (as Heima) and Vicuna-7B~\citep{zheng2023vicuna} (as interpreter) are adopted.
We train LLaVA-Next-Vicuna-7B on \texttt{LLaVA-CoT-100k} through LoRA to capture reasoning capability, and then perform our method with thinking tokens with this model family.

\textbf{Evaluation.  }
We adopt multiple challenging zero-shot benchmarks to verify the effectiveness of our proposed method, including MMStar~\citep{mmstar_dataset}, MMBench V1.1~\citep{mmbench}, MMVet~\citep{mmvet}, MathVista~\citep{mathvista}, AI2D~\citep{ai2d}, and HallusionBench~\citep{hallusionbench}. 
MMStar, MMBench, and MMVet evaluate general visual question-answering capabilities, while MathVista and AI2D assess mathematical and scientific reasoning. HallusionBench, in contrast, targets language hallucinations and visual illusions.
We use the VLMEvalKit~\citep{duan2024vlmevalkit} as the evaluation pipeline to ensure a fair comparison.
We reproduce the evaluation results of LLaVA-CoT to get the number of generated tokens.
GPT-4o~\citep{achiam2023gpt4_tech_report} is adopted for evaluation on the MMVet and MathVista datasets, while exact match evaluation is applied to other datasets using VLMEvalKit.
For Heima, we split the \texttt{LLaVA-CoT-100k} dataset for train and test separately.
We evaluate fine-tuned interpreters on test set which contain 4300 samples with metrics including BLEU-4~\citep{papineni2002bleu}, METEOR~\citep{banerjee2005meteor}, ROUGE~\citep{lin2004rouge}, BERTScore~\citep{zhang2019bertscore}, and similarity analysis from GPT-4o.




\begin{table*}[t]
\centering
\caption{
Results with LLaVA model family.  LoRA is used to train the  model for CoT improvements.} 
\vspace{-2mm}
\resizebox{0.99\linewidth}{!}{
\begin{tabular}{l|cccccc|c}
\toprule
Datasets & MMSar & MMBench & MMVet & MathVista & AI2D  & Hallusion & Avg.           \\
\midrule
\multirow{2}{*}{Model} 
    & Acc. & Acc. & Acc. & Acc. & Acc. & Acc. & \multirow{2}{*}{Acc.} \\
    & (\# Token) & (\# Token) & (\# Token) & (\# Token) & (\# Token) & (\# Token) & \\
\midrule
\begin{tabular}[c]{@{}l@{}}LLaVA-Next\\-Vicuna-7B \end{tabular} & 37.7 (2.5)  & 65.6 (2.0)  & 33.4 (143.9)  & 30.2 (93.7)     & 67.0 (2.0) & 32.3  (69.3)    & 44.4 \\
\midrule
\begin{tabular}[c]{@{}l@{}}LLaVA-Next\\-Vicuna-7B (CoT) \end{tabular} & 46.5 (175.9)  & 71.5 (155.3)   & 47.5 (230.5) & 41.8 (190.6)  & 77.3 (165.7) & 45.1  (149.8)    & 55.0 \\
\midrule
\midrule
\textbf{Heima} & 44.6 (12.8)  & 73.5 (12.5)   & 43.4 (68.9) & 40.6 (12.7)      & 77.1 (12.6) & 43.3   (15.7)   & 53.8 \\
\bottomrule
\end{tabular}
}
\vspace{-3mm}
\label{tab:results_llava}
\end{table*}

\subsection{Main Results}

We first provide main results for Heima in Table~\ref{tab:results_main}.
We compare our method with original Llama3.2-11B-Vision-Instruct and the LLaVA-CoT on 6 datasets for zero-shot evaluation.
Heima outperforms Llama3.2-11B-Vision-Instruct model with large improvements in average accuracy, while using fewer tokens, particularly on benchmarks such as MMBench, AI2D, and Hallusion with much higher accuracy and fewer tokens.
Compared with baseline LLaVA-CoT, Heima retains most of the model's performance while using as little as 6\% of the tokens on certain datasets.
Notably, on MMBench, Heima achieves better accuracy than the baseline LLaVA-CoT.
Furthermore, to demonstrate the effectiveness of progressive distillation, we show the accuracy results using one-shot distillation to distill all CoT stages in one shot.
Results achieved with non-progressive distillation indicate worse performance, confirming the effectiveness of the progressive distillation in our framework.
Additionally, the accuracy results without the recovering stage highlight its necessity, as they demonstrate a noticeable decline in performance compared to that with the recovering stage after completing the distillation of all CoT stages.
We further present detailed accuracy results for various reasoning tasks of MMStar in Table~\ref{tab:results_mmstar} of Appendix~\ref{sec:supp_additional_results}.
Heima outperforms Llama3.2-11B on both instance reasoning (IR) and logical reasoning (LR) tasks while using less than 10\% of tokens, and it preserves the majority of its reasoning capabilities for mathematical problems through progressive distillation.



Meanwhile, we provide additional results with LLaVA-Next-Vicuna-7B in Table~\ref{tab:results_llava} to verify the generalization of our framework.
Our method achieves better performance than non-CoT model (i.e., the original LLaVA-Next-Vicuna-7B). Compared with the LoRA fine-tuned CoT model, our method achieves comparable accuracy with significantly fewer generated tokens (as little as 6\%).
The consistent performance on different models architectures demonstrates the effectiveness, efficiency, and generalization of our method.

\begin{figure*}[t]
  \centering
  \includegraphics[width=0.99\linewidth]{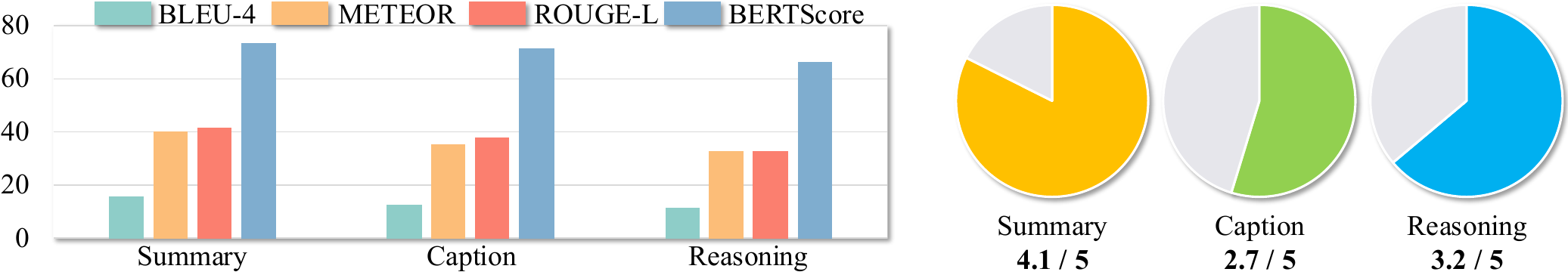}
  \vspace{-1mm}
  \caption{
  \textbf{Left}: results of BLEU-4, METEOR, GROUGE-L, and BERTScore for 3 interpreters.
  \textbf{Right}: results of evaluation by GPT-4o for evaluating the average similarity score (1-5) between the reconstructed reasoning processes from thinking tokens and the original textual CoTs.
  }
  \label{fig:results_decoder_eval_metrics_gpt4o}
  \vspace{-1mm}
\end{figure*}

\begin{figure*}[t]
  \centering
  \includegraphics[width=0.99\linewidth]{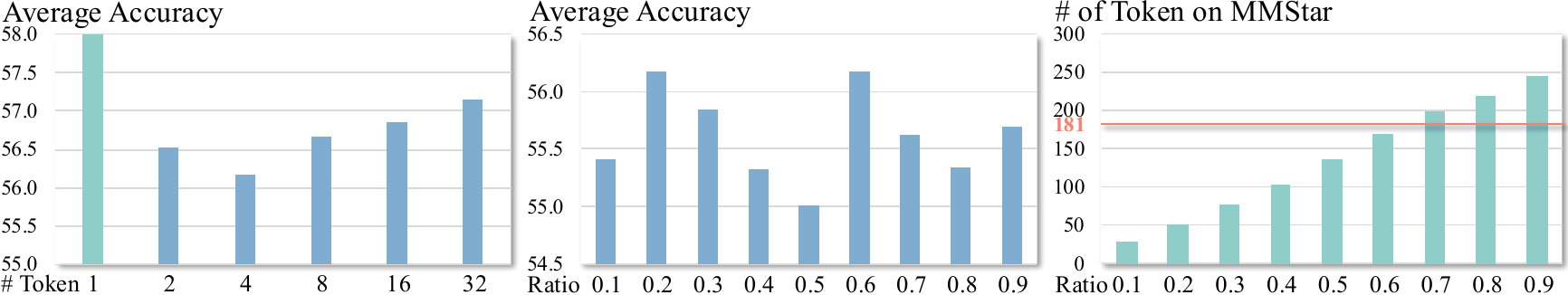}
  \vspace{-1mm}
  \caption{
  \textbf{Left}: ablation study of zero-shot performance on 6 datasets for different number of thinking tokens for each CoT.
  \textbf{Mid}: ablation study of average accuracy on 6 datasets for varying retention ratios of thinking tokens relative to original CoT.
  \textbf{Right}: ablation study for the number of generated tokens on MMStar with varying retention ratios of thinking tokens relative to the original CoT. Baseline (i.e., LLaVA-CoT) generates 181 tokens in average.
  }
  \label{fig:ablation_study}
\end{figure*}



\subsection{Interpretability Analysis}

To quantify the information gap  by the compression as illustrated in Section~\ref{sec:theory_analysis}, we evaluate the similarity between interpreter-reconstructed reasoning sentences and the ground-truth textual CoTs.
We provide results of 4 evaluation metrics in the left side of Figure~\ref{fig:results_decoder_eval_metrics_gpt4o} with details in 
Table~\ref{tab:supp_results_detailed_decoder_eval_metrics} of Appendix~\ref{sec:supp_additional_results}.
We observe that the reconstruction is most successful for summary stage, followed by caption stage, and then reasoning stage. 
In addition, we use GPT-4o to evaluate the similarity between reconstructed reasoning and original CoTs using a 5-point ranking scale (higher is better), and the results are shown in the right side of Figure~\ref{fig:results_decoder_eval_metrics_gpt4o}.
Prompts for GPT-4o are included in Appendix~\ref{sec:supp_gpt4o_prompts}.
We average the rank of all samples in one stage to estimate the similarity score, and results verifies that all stages are effectively reconstructed.

Meanwhile, we provide evaluation results for interpreter with the LLaVA model family in Table~\ref{tab:supp_results_llava_decoder_eval_metrics} of Appendix~\ref{sec:supp_additional_results}. 
Results are consistent with those in Figure~\ref{fig:results_decoder_eval_metrics_gpt4o} from Llama3 model family, further demonstrating the effectiveness and generalization of our method.

The results discussed above demonstrate that the information gap introduced by Heima between verbose textual CoTs and compact thinking tokens is minimal, thereby validating the theoretical analysis in Section~\ref{sec:theory_analysis}.
More importantly, the reasoning capability of the compressed representation is well preserved, indicating that the retained non-trivial mutual information is sufficient for sustaining effective reasoning in latent space.
In particular, the example in Figure~\ref{fig:introduction_whole_pipeline} highlights that the interpreter successfully reconstructs textual reasoning progress that capture the key insights of visual information even when no visual input is provided.
This finding confirms that thinking tokens preserve multimodal signals for reasoning, underscoring both the effectiveness of the compression with Heima and the robustness of reasoning in latent space.



\subsection{Ablation Study}

We provide results with different numbers of thinking tokens adopted for the distillation of one CoT stage in left side of Figure~\ref{fig:ablation_study} with detailed results in Table~\ref{tab:supp_results_detailed_ablation_num_tokens} of Appendix~\ref{sec:supp_additional_results}.
Results show that the single thinking token for encoding CoT stage achieves the best performance.

We further ablate for adaptive distillation by using different retention ratio of the original CoT token length, and the corresponding results are in middle of Figure~\ref{fig:ablation_study} with detailed results in Table~\ref{tab:supp_results_detailed_ablation_ratio_tokens} of Appendix~\ref{sec:supp_additional_results}.
Across retention ratios ranging from 10\% to 90\%, the accuracy exhibits irregular fluctuations without a discernible trend, reflecting the inherently unpredictable relationship between retention ratio and accuracy.
Also, as shown in right side of Figure~\ref{fig:ablation_study}, the average number of generated token keeps increasing as retention ratio becomes larger.
When retention ratio reaches 70\%, the average number of generated tokens exceeds that of baseline model, indicating the adaptive distillation is not effective for the compression of the reasoning progress. 

In addition, we conduct the ablation study on the number of interpreters to examine whether it is necessary to adopt distinct interpreters for each CoT stage. The results are reported in Table~\ref{tab:supp_results_ablation_number_of_decoder} in Appendix~\ref{sec:supp_additional_ablation_num_of_decoders}.
The ablation results reveal that distinct interpreters are crucial for summary and caption reconstruction.
We emphasize that interpreters are not integrated into the efficient reasoning procedure of Heima, but are instead utilized exclusively to interpret the reasoning in latent space.

\section{Conclusion}

We introduced Heima, a framework for accelerating reasoning in MLLMs through CoT compression.
Heima distills each CoT into a compact thinking token and is supported by an information-theoretic analysis that quantifies the compression-induced information gap.
To empirically examine this gap, we further design the interpreters guided by explanatory prompts to reconstruct reasoning progresses from thinking tokens.
Extensive experiments demonstrate that Heima achieves comparable or even superior zero-shot accuracy with significantly fewer tokens, highlighting both its efficiency and robustness.
Moreover, the successful reconstruction of reasoning processes confirms that the information gap is minimal and reasoning capability is preserved.
Looking ahead, we will extend Heima to larger models.

\section*{Impact Statement}


This paper presents work whose goal is to advance the field of Machine
Learning. There are many potential societal consequences of our work, none
which we feel must be specifically highlighted here.

\bibliography{reference}

@article{liu2024llava,
  title={Visual instruction tuning},
  author={Liu, Haotian and Li, Chunyuan and Wu, Qingyang and Lee, Yong Jae},
  journal={Advances in neural information processing systems},
  volume={36},
  year={2024}
}

@misc{liu2024llavanext,
    title={LLaVA-NeXT: Improved reasoning, OCR, and world knowledge},
    url={https://llava-vl.github.io/blog/2024-01-30-llava-next/},
    author={Liu, Haotian and Li, Chunyuan and Li, Yuheng and Li, Bo and Zhang, Yuanhan and Shen, Sheng and Lee, Yong Jae},
    month={January},
    year={2024}
}

@misc{zheng2023vicuna,
      title={Judging LLM-as-a-judge with MT-Bench and Chatbot Arena},
      author={Lianmin Zheng and Wei-Lin Chiang and Ying Sheng and Siyuan Zhuang and Zhanghao Wu and Yonghao Zhuang and Zi Lin and Zhuohan Li and Dacheng Li and Eric. P Xing and Hao Zhang and Joseph E. Gonzalez and Ion Stoica},
      year={2023},
      eprint={2306.05685},
      archivePrefix={arXiv},
      primaryClass={cs.CL}
}

@article{qwen-VL,
  title={Qwen-VL: A Versatile Vision-Language Model for Understanding, Localization, Text Reading, and Beyond},
  author={Bai, Jinze and Bai, Shuai and Yang, Shusheng and Wang, Shijie and Tan, Sinan and Wang, Peng and Lin, Junyang and Zhou, Chang and Zhou, Jingren},
  journal={arXiv preprint arXiv:2308.12966},
  year={2023}
}

@article{achiam2023gpt4_tech_report,
  title={Gpt-4 technical report},
  author={Achiam, Josh and Adler, Steven and Agarwal, Sandhini and Ahmad, Lama and Akkaya, Ilge and Aleman, Florencia Leoni and Almeida, Diogo and Altenschmidt, Janko and Altman, Sam and Anadkat, Shyamal and others},
  journal={arXiv preprint arXiv:2303.08774},
  year={2023}
}

@article{wei2022original_cot,
  title={Chain-of-thought prompting elicits reasoning in large language models},
  author={Wei, Jason and Wang, Xuezhi and Schuurmans, Dale and Bosma, Maarten and Xia, Fei and Chi, Ed and Le, Quoc V and Zhou, Denny and others},
  journal={Advances in neural information processing systems},
  volume={35},
  pages={24824--24837},
  year={2022}
}

@article{xu2024llavacot,
  title={LLaVA-o1: Let Vision Language Models Reason Step-by-Step},
  author={Xu, Guowei and Jin, Peng and Hao, Li and Song, Yibing and Sun, Lichao and Yuan, Li},
  journal={arXiv preprint arXiv:2411.10440},
  year={2024}
}

@article{hao2024meta_compress_cot,
  title={Training large language models to reason in a continuous latent space},
  author={Hao, Shibo and Sukhbaatar, Sainbayar and Su, DiJia and Li, Xian and Hu, Zhiting and Weston, Jason and Tian, Yuandong},
  journal={arXiv preprint arXiv:2412.06769},
  year={2024}
}

@article{deng2024compress_cot,
  title={From explicit cot to implicit cot: Learning to internalize cot step by step},
  author={Deng, Yuntian and Choi, Yejin and Shieber, Stuart},
  journal={arXiv preprint arXiv:2405.14838},
  year={2024}
}

@article{radford2019gpt2,
  title={Rewon child, david luan, dario amodei, and ilya sutskever. 2019},
  author={Radford, Alec and Wu, Jeffrey},
  journal={Language models are unsupervised multitask learners. OpenAI blog},
  volume={1},
  number={8},
  pages={9},
  year={2019}
}

@inproceedings{lai2024lisa,
  title={Lisa: Reasoning segmentation via large language model},
  author={Lai, Xin and Tian, Zhuotao and Chen, Yukang and Li, Yanwei and Yuan, Yuhui and Liu, Shu and Jia, Jiaya},
  booktitle={Proceedings of the IEEE/CVF Conference on Computer Vision and Pattern Recognition},
  pages={9579--9589},
  year={2024}
}

@article{hu2021lora,
  title={Lora: Low-rank adaptation of large language models},
  author={Hu, Edward J and Shen, Yelong and Wallis, Phillip and Allen-Zhu, Zeyuan and Li, Yuanzhi and Wang, Shean and Wang, Lu and Chen, Weizhu},
  journal={arXiv preprint arXiv:2106.09685},
  year={2021}
}

@article{khot2022decomposed,
  title={Decomposed prompting: A modular approach for solving complex tasks},
  author={Khot, Tushar and Trivedi, Harsh and Finlayson, Matthew and Fu, Yao and Richardson, Kyle and Clark, Peter and Sabharwal, Ashish},
  journal={arXiv preprint arXiv:2210.02406},
  year={2022}
}

@article{zhou2022least,
  title={Least-to-most prompting enables complex reasoning in large language models},
  author={Zhou, Denny and Sch{\"a}rli, Nathanael and Hou, Le and Wei, Jason and Scales, Nathan and Wang, Xuezhi and Schuurmans, Dale and Cui, Claire and Bousquet, Olivier and Le, Quoc and others},
  journal={arXiv preprint arXiv:2205.10625},
  year={2022}
}

@article{yue2023mammoth,
  title={Mammoth: Building math generalist models through hybrid instruction tuning},
  author={Yue, Xiang and Qu, Xingwei and Zhang, Ge and Fu, Yao and Huang, Wenhao and Sun, Huan and Su, Yu and Chen, Wenhu},
  journal={arXiv preprint arXiv:2309.05653},
  year={2023}
}

@article{yu2023metamath,
  title={Metamath: Bootstrap your own mathematical questions for large language models},
  author={Yu, Longhui and Jiang, Weisen and Shi, Han and Yu, Jincheng and Liu, Zhengying and Zhang, Yu and Kwok, James T and Li, Zhenguo and Weller, Adrian and Liu, Weiyang},
  journal={arXiv preprint arXiv:2309.12284},
  year={2023}
}

@article{wang2023math,
  title={Math-shepherd: A label-free step-by-step verifier for llms in mathematical reasoning},
  author={Wang, Peiyi and Li, Lei and Shao, Zhihong and Xu, RX and Dai, Damai and Li, Yifei and Chen, Deli and Wu, Y and Sui, Zhifang},
  journal={arXiv preprint arXiv:2312.08935},
  year={2023}
}

@article{shao2024deepseekmath,
  title={Deepseekmath: Pushing the limits of mathematical reasoning in open language models},
  author={Shao, Zhihong and Wang, Peiyi and Zhu, Qihao and Xu, Runxin and Song, Junxiao and Bi, Xiao and Zhang, Haowei and Zhang, Mingchuan and Li, YK and Wu, Y and others},
  journal={arXiv preprint arXiv:2402.03300},
  year={2024}
}

@article{liu2024deepseekv3,
  title={Deepseek-v3 technical report},
  author={Liu, Aixin and Feng, Bei and Xue, Bing and Wang, Bingxuan and Wu, Bochao and Lu, Chengda and Zhao, Chenggang and Deng, Chengqi and Zhang, Chenyu and Ruan, Chong and others},
  journal={arXiv preprint arXiv:2412.19437},
  year={2024}
}

@article{feng2024towards,
  title={Towards revealing the mystery behind chain of thought: a theoretical perspective},
  author={Feng, Guhao and Zhang, Bohang and Gu, Yuntian and Ye, Haotian and He, Di and Wang, Liwei},
  journal={Advances in Neural Information Processing Systems},
  volume={36},
  year={2024}
}

@article{merrill2023expresssive,
  title={The expresssive power of transformers with chain of thought},
  author={Merrill, William and Sabharwal, Ashish},
  journal={arXiv preprint arXiv:2310.07923},
  year={2023}
}

@article{xie2024self,
  title={Self-evaluation guided beam search for reasoning},
  author={Xie, Yuxi and Kawaguchi, Kenji and Zhao, Yiran and Zhao, James Xu and Kan, Min-Yen and He, Junxian and Xie, Michael},
  journal={Advances in Neural Information Processing Systems},
  volume={36},
  year={2024}
}

@article{gandhi2024stream,
  title={Stream of Search (SoS): Learning to Search in Language},
  author={Gandhi, Kanishk and Lee, Denise and Grand, Gabriel and Liu, Muxin and Cheng, Winson and Sharma, Archit and Goodman, Noah D},
  journal={arXiv preprint arXiv:2404.03683},
  year={2024}
}

@article{su2024dualformer,
  title={Dualformer: Controllable fast and slow thinking by learning with randomized reasoning traces},
  author={Su, DiJia and Sukhbaatar, Sainbayar and Rabbat, Michael and Tian, Yuandong and Zheng, Qinqing},
  journal={arXiv preprint arXiv:2410.09918},
  year={2024}
}

@article{yang2024large,
  title={Do Large Language Models Latently Perform Multi-Hop Reasoning?},
  author={Yang, Sohee and Gribovskaya, Elena and Kassner, Nora and Geva, Mor and Riedel, Sebastian},
  journal={arXiv preprint arXiv:2402.16837},
  year={2024}
}

@article{biran2024hopping,
  title={Hopping Too Late: Exploring the Limitations of Large Language Models on Multi-Hop Queries},
  author={Biran, Eden and Gottesman, Daniela and Yang, Sohee and Geva, Mor and Globerson, Amir},
  journal={arXiv preprint arXiv:2406.12775},
  year={2024}
}

@article{pi2023detgpt,
  title={Detgpt: Detect what you need via reasoning},
  author={Pi, Renjie and Gao, Jiahui and Diao, Shizhe and Pan, Rui and Dong, Hanze and Zhang, Jipeng and Yao, Lewei and Han, Jianhua and Xu, Hang and Kong, Lingpeng and others},
  journal={arXiv preprint arXiv:2305.14167},
  year={2023}
}

@article{deng2024motion,
  title={Motion-Grounded Video Reasoning: Understanding and Perceiving Motion at Pixel Level},
  author={Deng, Andong and Chen, Tongjia and Yu, Shoubin and Yang, Taojiannan and Spencer, Lincoln and Tian, Yapeng and Mian, Ajmal Saeed and Bansal, Mohit and Chen, Chen},
  journal={arXiv preprint arXiv:2411.09921},
  year={2024}
}

@article{yan2024visa,
  title={VISA: Reasoning Video Object Segmentation via Large Language Models},
  author={Yan, Cilin and Wang, Haochen and Yan, Shilin and Jiang, Xiaolong and Hu, Yao and Kang, Guoliang and Xie, Weidi and Gavves, Efstratios},
  journal={arXiv preprint arXiv:2407.11325},
  year={2024}
}

@article{llama32vision,
  title={Llama 3.2: Revolutionizing edge AI and vision with open, customizable models},
  author={Meta},
  url={https://ai.meta.com/blog/llama-3-2-connect-2024-vision-edge-mobile-devices/},
journal={blog},
  year={2024}
}

@article{llama31,
  title={Introducing Llama 3.1: Our most capable models to date},
  author={Meta},
  url={https://ai.meta.com/blog/meta-llama-3-1/},
    journal={blog},
  year={2024}
}

@article{torchtune,
  title = {torchtune: PyTorch's finetuning library},
  author = {Meta},
  url = {https//github.com/pytorch/torchtune},
  license = {BSD-3-Clause},
  month = apr,
  journal={software},
  year = {2024}
}

@article{mmstar_dataset,
  title={Are We on the Right Way for Evaluating Large Vision-Language Models?},
  author={Chen, Lin and Li, Jinsong and Dong, Xiaoyi and Zhang, Pan and Zang, Yuhang and Chen, Zehui and Duan, Haodong and Wang, Jiaqi and Qiao, Yu and Lin, Dahua and others},
  journal={arXiv preprint arXiv:2403.20330},
  year={2024}
}

@inproceedings{mmbench,
  title={Mmbench: Is your multi-modal model an all-around player?},
  author={Liu, Yuan and Duan, Haodong and Zhang, Yuanhan and Li, Bo and Zhang, Songyang and Zhao, Wangbo and Yuan, Yike and Wang, Jiaqi and He, Conghui and Liu, Ziwei and others},
  booktitle={European conference on computer vision},
  pages={216--233},
  year={2025},
  organization={Springer}
}

@inproceedings{mmvet,
  title={Mm-vet: Evaluating large multimodal models for integrated capabilities},
  author={Yu, Weihao and Yang, Zhengyuan and Li, Linjie and Wang, Jianfeng and Lin, Kevin and Liu, Zicheng and Wang, Xinchao and Wang, Lijuan},
  booktitle={International conference on machine learning},
  year={2024},
  organization={PMLR}
}

@inproceedings{mathvista,
  author    = {Lu, Pan and Bansal, Hritik and Xia, Tony and Liu, Jiacheng and Li, Chunyuan and Hajishirzi, Hannaneh and Cheng, Hao and Chang, Kai-Wei and Galley, Michel and Gao, Jianfeng},
  title     = {MathVista: Evaluating Mathematical Reasoning of Foundation Models in Visual Contexts},
  booktitle={International Conference on Learning Representations (ICLR)},
  year      = {2024}
}

@article{ai2d,
  title={AI2D-RST: A multimodal corpus of 1000 primary school science diagrams},
  author={Hiippala, Tuomo and Alikhani, Malihe and Haverinen, Jonas and Kalliokoski, Timo and Logacheva, Evanfiya and Orekhova, Serafina and Tuomainen, Aino and Stone, Matthew and Bateman, John A},
  journal={Language Resources and Evaluation},
  volume={55},
  pages={661--688},
  year={2021},
  publisher={Springer}
}

@InProceedings{hallusionbench,
    author    = {Guan, Tianrui and Liu, Fuxiao and Wu, Xiyang and Xian, Ruiqi and Li, Zongxia and Liu, Xiaoyu and Wang, Xijun and Chen, Lichang and Huang, Furong and Yacoob, Yaser and Manocha, Dinesh and Zhou, Tianyi},
    title     = {HallusionBench: An Advanced Diagnostic Suite for Entangled Language Hallucination and Visual Illusion in Large Vision-Language Models},
    booktitle = {Proceedings of the IEEE/CVF Conference on Computer Vision and Pattern Recognition (CVPR)},
    month     = {June},
    year      = {2024},
    pages     = {14375-14385}
}

@inproceedings{duan2024vlmevalkit,
  title={Vlmevalkit: An open-source toolkit for evaluating large multi-modality models},
  author={Duan, Haodong and Yang, Junming and Qiao, Yuxuan and Fang, Xinyu and Chen, Lin and Liu, Yuan and Dong, Xiaoyi and Zang, Yuhang and Zhang, Pan and Wang, Jiaqi and others},
  booktitle={Proceedings of the 32nd ACM International Conference on Multimedia},
  pages={11198--11201},
  year={2024}
}

@article{zhang2019bertscore,  
  title={BERTScore: Evaluating text generation with BERT},  
  author={Zhang, Tianyi and Kishore, Varsha and Wu, Felix and Weinberger, Kilian Q and Artzi, Yoav},  
  journal={arXiv preprint arXiv:1904.09675},  
  year={2019}  
}

@inproceedings{papineni2002bleu,
  title={Bleu: a method for automatic evaluation of machine translation},
  author={Papineni, Kishore and Roukos, Salim and Ward, Todd and Zhu, Wei-Jing},
  booktitle={ACL},
  year={2002}
}

@inproceedings{banerjee2005meteor,
  title={METEOR: An automatic metric for MT evaluation with improved correlation with human judgments},
  author={Banerjee, Satanjeev and Lavie, Alon},
  booktitle={Proceedings of the acl workshop on intrinsic and extrinsic evaluation measures for machine translation and/or summarization},
  pages={65--72},
  year={2005}
}

@inproceedings{lin2004rouge,  
  title={ROUGE: A package for automatic evaluation of summaries},  
  author={Lin, Chin-Yew},  
  booktitle={Text Summarization Branches Out},  
  pages={74--81},  
  year={2004}  
}

@article{cheng2024compressed,
  title={Compressed chain of thought: Efficient reasoning through dense representations},
  author={Cheng, Jeffrey and Van Durme, Benjamin},
  journal={arXiv preprint arXiv:2412.13171},
  year={2024}
}

@article{ning2023skeleton,
  title={Skeleton-of-thought: Large language models can do parallel decoding},
  author={Ning, Xuefei and Lin, Zinan and Zhou, Zixuan and Wang, Zifu and Yang, Huazhong and Wang, Yu},
  journal={Proceedings ENLSP-III},
  year={2023}
}

@article{kou2024cllms,
  title={Cllms: Consistency large language models},
  author={Kou, Siqi and Hu, Lanxiang and He, Zhezhi and Deng, Zhijie and Zhang, Hao},
  journal={arXiv preprint arXiv:2403.00835},
  year={2024}
}

@article{li2024eagle,
  title={Eagle-2: Faster inference of language models with dynamic draft trees},
  author={Li, Yuhui and Wei, Fangyun and Zhang, Chao and Zhang, Hongyang},
  journal={arXiv preprint arXiv:2406.16858},
  year={2024}
}

@article{zhang2023fast,
  title={Fast Chain-of-Thought: A Glance of Future from Parallel Decoding Leads to Answers Faster},
  author={Zhang, Hongxuan and Liu, Zhining and Zhao, Yao and Zheng, Jiaqi and Zhuang, Chenyi and Gu, Jinjie and Chen, Guihai},
  journal={arXiv preprint arXiv:2311.08263},
  year={2023}
}

@article{chevalier2023adapting,
  title={Adapting language models to compress contexts},
  author={Chevalier, Alexis and Wettig, Alexander and Ajith, Anirudh and Chen, Danqi},
  journal={arXiv preprint arXiv:2305.14788},
  year={2023}
}

@inproceedings{
  ge2024incontext,
  title={In-context Autoencoder for Context Compression in a Large Language Model},
  author={Tao Ge and Hu Jing and Lei Wang and Xun Wang and Si-Qing Chen and Furu Wei},
  booktitle={The Twelfth International Conference on Learning Representations},
  year={2024},
  url={https://openreview.net/forum?id=uREj4ZuGJE}
}

@article{qin2023nugget,
  title={Nugget 2D: Dynamic Contextual Compression for Scaling Decoder-only Language Models},
  author={Qin, Guanghui and Rosset, Corby and Chau, Ethan C and Rao, Nikhil and Van Durme, Benjamin},
  journal={arXiv preprint arXiv:2310.02409},
  year={2023}
}

@article{liu2023cachegen,
  title={Cachegen: Fast context loading for language model applications},
  author={Liu, Yuhan and Li, Hanchen and Du, Kuntai and Yao, Jiayi and Cheng, Yihua and Huang, Yuyang and Lu, Shan and Maire, Michael and Hoffmann, Henry and Holtzman, Ari and others},
  journal={arXiv preprint arXiv:2310.07240},
  year={2023}
}

@article{munkhdalai2024leave,
  title={Leave no context behind: Efficient infinite context transformers with infini-attention},
  author={Munkhdalai, Tsendsuren and Faruqui, Manaal and Gopal, Siddharth},
  journal={arXiv preprint arXiv:2404.07143},
  year={2024}
}

@inproceedings{
huang2024compression,
title={Compression Represents Intelligence Linearly},
author={Yuzhen Huang and Jinghan Zhang and Zifei Shan and Junxian He},
booktitle={First Conference on Language Modeling},
year={2024},
url={https://openreview.net/forum?id=SHMj84U5SH}
}

@inproceedings{deletang2024language,
  author       = {Gr{\'{e}}goire Del{\'{e}}tang and
                  Anian Ruoss and
                  Paul{-}Ambroise Duquenne and
                  Elliot Catt and
                  Tim Genewein and
                  Christopher Mattern and
                  Jordi Grau{-}Moya and
                  Li Kevin Wenliang and
                  Matthew Aitchison and
                  Laurent Orseau and
                  Marcus Hutter and
                  Joel Veness},
  title        = {Language Modeling Is Compression},
  booktitle    = {{ICLR}},
  year         = {2024}
}

@book{mackay2003information,
  title={Information theory, inference and learning algorithms},
  author={MacKay, David JC},
  year={2003},
  publisher={Cambridge university press}
}

@inproceedings{zhou2025geometry,
title     = {The Geometry of Reasoning: Flowing Logics in Representation Space},
author    = {Zhou, Yufa and Wang, Yixiao and Yin, Xunjian and Zhou, Shuyan and Zhang, Anru R.},
booktitle = {The Fourteenth International Conference on Learning Representations},
year      = {2026},
url       = {https://openreview.net/forum?id=ixr5Pcabq7}
}

@InProceedings{shen2023deepmad,
    author    = {Shen, Xuan and Wang, Yaohua and Lin, Ming and Huang, Yilun and Tang, Hao and Sun, Xiuyu and Wang, Yanzhi},
    title     = {DeepMAD: Mathematical Architecture Design for Deep Convolutional Neural Network},
    booktitle = {Proceedings of the IEEE/CVF Conference on Computer Vision and Pattern Recognition (CVPR)},
    month     = {June},
    year      = {2023},
    pages     = {6163-6173}
}

@article{shen2024agile, title={Agile-Quant: Activation-Guided Quantization for Faster Inference of LLMs on the Edge}, volume={38}, url={https://ojs.aaai.org/index.php/AAAI/article/view/29860}, DOI={10.1609/aaai.v38i17.29860}, number={17}, journal={Proceedings of the AAAI Conference on Artificial Intelligence}, author={Shen, Xuan and Dong, Peiyan and Lu, Lei and Kong, Zhenglun and Li, Zhengang and Lin, Ming and Wu, Chao and Wang, Yanzhi}, year={2024}, month={Mar.}, pages={18944-18951} }

@article{shen2025lazydit, title={LazyDiT: Lazy Learning for the Acceleration of Diffusion Transformers}, volume={39}, url={https://ojs.aaai.org/index.php/AAAI/article/view/34248}, DOI={10.1609/aaai.v39i19.34248}, number={19}, journal={Proceedings of the AAAI Conference on Artificial Intelligence}, author={Shen, Xuan and Song, Zhao and Zhou, Yufa and Chen, Bo and Li, Yanyu and Gong, Yifan and Zhang, Kai and Tan, Hao and Kuen, Jason and Ding, Henghui and Shu, Zhihao and Niu, Wei and Zhao, Pu and Wang, Yanzhi and Gu, Jiuxiang}, year={2025}, month={Apr.}, pages={20409-20417} }

@article{shen2025numerical, title={Numerical Pruning for Efficient Autoregressive Models}, volume={39}, url={https://ojs.aaai.org/index.php/AAAI/article/view/34249}, DOI={10.1609/aaai.v39i19.34249}, number={19}, journal={Proceedings of the AAAI Conference on Artificial Intelligence}, author={Shen, Xuan and Song, Zhao and Zhou, Yufa and Chen, Bo and Liu, Jing and Zhang, Ruiyi and Rossi, Ryan A. and Tan, Hao and Yu, Tong and Chen, Xiang and Zhou, Yufan and Sun, Tong and Zhao, Pu and Wang, Yanzhi and Gu, Jiuxiang}, year={2025}, month={Apr.}, pages={20418-20426} }

@inproceedings{add-nguyen-etal-2025-survey,
    title = "A Survey on Small Language Models",
    author = "Nguyen, Chien Van  and
      Shen, Xuan  and
      Aponte, Ryan  and
      Xia, Yu  and
      Basu, Samyadeep  and
      Hu, Zhengmian  and
      Chen, Jian  and
      Parmar, Mihir  and
      Kunapuli, Sasidhar  and
      Barrow, Joe  and
      Wu, Junda  and
      Singh, Ashish  and
      Wang, Yu  and
      Gu, Jiuxiang  and
      K. Ahmed, Nesreen  and
      Lipka, Nedim  and
      Zhang, Ruiyi  and
      Chen, Xiang  and
      Yu, Tong  and
      Kim, Sungchul  and
      Deilamsalehy, Hanieh  and
      Park, Namyong  and
      Rimer, Michael  and
      Zhang, Zhehao  and
      Yang, Huanrui  and
      Mathur, Puneet  and
      Wu, Gang  and
      Dernoncourt, Franck  and
      Rossi, Ryan A.  and
      Nguyen, Thien Huu",
    editor = "Angelova, Galia  and
      Kunilovskaya, Maria  and
      Escribe, Marie  and
      Mitkov, Ruslan",
    booktitle = "Proceedings of the 15th International Conference on Recent Advances in Natural Language Processing - Natural Language Processing in the Generative AI Era",
    month = sep,
    year = "2025",
    address = "Varna, Bulgaria",
    publisher = "INCOMA Ltd., Shoumen, Bulgaria",
    url = "https://aclanthology.org/2025.ranlp-1.93/",
    pages = "807--821",
}

@INPROCEEDINGS{shen2025iccad,
  author={Shen, Xuan and Dong, Peiyan and Kong, Zhenglun and Gong, Yifan and Yang, Changdi and Han, Zhaoyang and Xie, Yanyue and Lu, Lei and Lyu, Cheng and Wu, Chao and Wang, Yanzhi and Zhao, Pu},
  booktitle={2025 IEEE/ACM International Conference On Computer Aided Design (ICCAD)}, 
  title={Squat: Quant Small Language Models on the Edge}, 
  year={2025},
  volume={},
  number={},
  pages={1-9},
  doi={10.1109/ICCAD66269.2025.11240685}}

@inproceedings{add-zhao-etal-2024-pruning,
    title = "Pruning Foundation Models for High Accuracy without Retraining",
    author = "Zhao, Pu  and
      Sun, Fei  and
      Shen, Xuan  and
      Yu, Pinrui  and
      Kong, Zhenglun  and
      Wang, Yanzhi  and
      Lin, Xue",
    editor = "Al-Onaizan, Yaser  and
      Bansal, Mohit  and
      Chen, Yun-Nung",
    booktitle = "Findings of the Association for Computational Linguistics: EMNLP 2024",
    month = nov,
    year = "2024",
    address = "Miami, Florida, USA",
    publisher = "Association for Computational Linguistics",
    url = "https://aclanthology.org/2024.findings-emnlp.566/",
    doi = "10.18653/v1/2024.findings-emnlp.566",
    pages = "9681--9694",
}

@inproceedings{shen2024search,
 author = {Shen, Xuan and Zhao, Pu and Gong, Yifan and Kong, Zhenglun and Zhan, Zheng and Wu, Yushu and Lin, Ming and Wu, Chao and Lin, Xue and Wang, Yanzhi},
 booktitle = {Advances in Neural Information Processing Systems},
 doi = {10.52202/079017-4421},
 editor = {A. Globerson and L. Mackey and D. Belgrave and A. Fan and U. Paquet and J. Tomczak and C. Zhang},
 pages = {139294--139315},
 publisher = {Curran Associates, Inc.},
 title = {Search for Efficient Large Language Models},
 url = {https://proceedings.neurips.cc/paper_files/paper/2024/file/fb64a43508e0cfe53ee6179ff31ea900-Paper-Conference.pdf},
 volume = {37},
 year = {2024}
}

@article{add-Liu-2025-toward, title={Toward Adaptive Large Language Models Structured Pruning via Hybrid-grained Weight Importance Assessment}, volume={39}, url={https://ojs.aaai.org/index.php/AAAI/article/view/34078}, DOI={10.1609/aaai.v39i18.34078}, number={18}, journal={Proceedings of the AAAI Conference on Artificial Intelligence}, author={Liu, Jun and Kong, Zhenglun and Zhao, Pu and Yang, Changdi and Shen, Xuan and Tang, Hao and Yuan, Geng and Niu, Wei and Zhang, Wenbin and Lin, Xue and Huang, Dong and Wang, Yanzhi}, year={2025}, month={Apr.}, pages={18879-18887} }

@ARTICLE{shen2024hota,
  author={Shen, Xuan and Han, Zhaoyang and Lu, Lei and Kong, Zhenglun and Dong, Peiyan and Li, Zhengang and Xie, Yanyue and Wu, Chao and Leeser, Miriam and Zhao, Pu and Lin, Xue and Wang, Yanzhi},
  journal={IEEE Transactions on Computer-Aided Design of Integrated Circuits and Systems}, 
  title={HotaQ: Hardware Oriented Token Adaptive Quantization for Large Language Models}, 
  year={2024},
  volume={},
  number={},
  pages={1-1},
  doi={10.1109/TCAD.2024.3487781}}

@InProceedings{shen2025quart,
    author    = {Shen, Xuan and Ma, Weize and Liu, Jing and Yang, Changdi and Ding, Rui and Wang, Quanyi and Ding, Henghui and Niu, Wei and Wang, Yanzhi and Zhao, Pu and Lin, Jun and Gu, Jiuxiang},
    title     = {QuartDepth: Post-Training Quantization for Real-Time Depth Estimation on the Edge},
    booktitle = {Proceedings of the IEEE/CVF Conference on Computer Vision and Pattern Recognition (CVPR)},
    month     = {June},
    year      = {2025},
    pages     = {11448-11460}
}

@INPROCEEDINGS{add-liu-rora,
  author={Liu, Jun and Kong, Zhenglun and Dong, Peiyan and Shen, Xuan and Zhao, Pu and Tang, Hao and Yuan, Geng and Niu, Wei and Zhang, Wenbin and Lin, Xue and Huang, Dong and Wang, Yanzhi},
  booktitle={ICASSP 2025 - 2025 IEEE International Conference on Acoustics, Speech and Signal Processing (ICASSP)}, 
  title={RoRA: Efficient Fine-Tuning of LLM with Reliability Optimization for Rank Adaptation}, 
  year={2025},
  volume={},
  number={},
  pages={1-5},
  doi={10.1109/ICASSP49660.2025.10889613}}

@inproceedings{
shen2025sparse,
title={Sparse Learning for State Space Models on Mobile},
author={Xuan Shen and Hangyu Zheng and Yifan Gong and Zhenglun Kong and Changdi Yang and Zheng Zhan and Yushu Wu and Xue Lin and Yanzhi Wang and Pu Zhao and Wei Niu},
booktitle={The Thirteenth International Conference on Learning Representations},
year={2025},
url={https://openreview.net/forum?id=t8KLjiFNwn}
}

@inproceedings{
shen2026fastcar,
title={Fastcar: Cache Attentive Replay for Fast Auto-Regressive Video Generation on the Edge},
author={Xuan Shen and Weize Ma and Yufa Zhou and Enhao Tang and Yanyue Xie and Zhengang Li and Yifan Gong and Quanyi Wang and Henghui Ding and Yiwei Wang and Pu Zhao and Jun Lin and Jiuxiang Gu},
booktitle={The Fourteenth International Conference on Learning Representations},
year={2026},
url={https://openreview.net/forum?id=9f3Nukn6BA}
}

@ARTICLE{add-liu-tsla,
  author={Liu, Jun and Kong, Zhenglun and Zhao, Pu and Zeng, Weihao and Tang, Hao and Shen, Xuan and Yang, Changdi and Zhang, Wenbin and Yuan, Geng and Niu, Wei and Lin, Xue and Wang, Yanzhi},
  journal={IEEE Transactions on Computer-Aided Design of Integrated Circuits and Systems}, 
  title={TSLA: A Task-Specific Learning Adaptation for Semantic Segmentation on Autonomous Vehicles Platform}, 
  year={2025},
  volume={44},
  number={4},
  pages={1406-1419},
  doi={10.1109/TCAD.2024.3491015}}

@article{shen2026oida, title={OIDA-QA: A Multimodal Benchmark for Analyzing the Opioid Industry Documents Archive}, volume={40}, url={https://ojs.aaai.org/index.php/AAAI/article/view/41273}, DOI={10.1609/aaai.v40i46.41273}, number={46}, journal={Proceedings of the AAAI Conference on Artificial Intelligence}, author={Shen, Xuan and Wingenroth, Brian and Wang, Zichao and Kuen, Jason and Zhu, Wanrong and Zhang, Ruiyi and Wang, Yiwei and Ma, Lichun and Liu, Anqi and Liu, Hongfu and Sun, Tong and Hawkins, Kevin S. and Tasker, Kate and Alexander, G. Caleb and Gu, Jiuxiang}, year={2026}, month={Mar.}, pages={39249-39258} }

@article{kong2025,
      title={Token Reduction Should Go Beyond Efficiency in Generative Models -- From Vision, Language to Multimodality},
      author={Zhenglun Kong and Yize Li and others},
      journal={arXiv preprint arXiv:2505.18227},
      year={2025}
}

@inproceedings{
zhan2024exploring,
title={Exploring Token Pruning in Vision State Space Models},
author={Zheng Zhan and Zhenglun Kong and Yifan Gong and others},
booktitle={NeurIPS},
year={2024},
url={https://openreview.net/forum?id=eWiGn0Fcdx}
}

@inproceedings{zhan-etal-2024-rethinking-token,
    title = {Rethinking Token Reduction for State Space Models},
    author = {Zhan, Zheng  and Wu, Yushu  and Kong, Zhenglun  and others},
    booktitle = {EMNLP},
    month = {nov},
    year = {2024},
    publisher = {ACL},
    pages = {1686--1697}
}

@article{zhao2025open,
  title={Open-Source Multimodal Moxin Models with Moxin-VLM and Moxin-VLA},
  author={Zhao, Pu and Akbari, Arash and Shen, Xuan and Kong, Zhenglun and Shen, Yixin and Chang, Sung-En and Rupprecht, Timothy and Lu, Lei and Nan, Enfu and Yang, Changdi and others},
  journal={arXiv preprint arXiv:2512.22208},
  year={2025}
}

@article{lin2025vote,
  title={Vote: vision-language-action optimization with trajectory ensemble voting},
  author={Lin, Juyi and Taherin, Amir and Akbari, Arash and Akbari, Arman and Lu, Lei and Chen, Guangyu and Padir, Taskin and Yang, Xiaomeng and Chen, Weiwei and Li, Yiqian and others},
  journal={arXiv preprint arXiv:2507.05116},
  year={2025}
}

@inproceedings{wang2025cautious,
  title={Cautious next token prediction},
  author={Wang, Yizhou and Zhang, Lingzhi and Bai, Yue and Chiu, Mang Tik and Hu, Zhengmian and Zhang, Mingyuan and Dong, Qihua and Yin, Yu and Amirghodsi, Sohrab and Fu, Yun},
  booktitle={Findings of the Association for Computational Linguistics: ACL 2025},
  pages={25685--25697},
  year={2025}
}
\bibliographystyle{icml2026}

\newpage
\appendix
\onecolumn

\setcounter{table}{0}
\renewcommand{\thetable}{A\arabic{table}}

\setcounter{figure}{0}
\renewcommand{\thefigure}{A\arabic{figure}}

\begin{center}
    {\bf \Large Appendix}
\end{center}

\section{Training Hyperparameters}\label{sec:supp_training_hyperparameters}

We provide the hyperparameters for the progressive distillation, additional recovering, and adaptive interpretation training in Table~\ref{tab:supp_training_hyperparameters_progressive_encoding}, Table~\ref{tab:supp_training_hyperparameters_additional_recovering}, and Table~\ref{tab:supp_training_hyperparameters_adaptive decoding}.

\begin{table}[h]
\centering
\caption{
Training hyperparameters for progressive distillation.
}
\label{tab:supp_training_hyperparameters_progressive_encoding}
\begin{tabular}{l|c}
\toprule
Parameter                & Value    \\
\midrule
Epoch                    & 1        \\
Batch Size               & 6        \\
Gradient Accumulation    & 1        \\
Optimizer                & AdamW    \\
Weight Decay             & 0.01     \\
Learning Rate            & 1e-4 \\
Learning Rate scheduler  & cosine   \\
Warmup                   & 100      \\
Clip Gradient Norm       & 1        \\
Activation Checkpointing & TRUE     \\
FSDP                     & TRUE     \\
Bfloat16                 & TRUE     \\
\bottomrule
\end{tabular}
\end{table}

\begin{table}[h]
\centering
\caption{
Training hyperparameters for additional recovering after progressive distillation.
}
\label{tab:supp_training_hyperparameters_additional_recovering}
\begin{tabular}{l|c}
\toprule
Parameter                & Value    \\
\midrule
Epoch                    & 1        \\
Batch Size               & 8        \\
Gradient Accumulation    & 1        \\
Optimizer                & AdamW    \\
Weight Decay             & 0.01     \\
Learning Rate            & 1e-5 \\
Learning Rate scheduler  & cosine   \\
Warmup                   & 100      \\
Clip Gradient Norm       & 1        \\
Activation Checkpointing & TRUE     \\
FSDP                     & TRUE     \\
Bfloat16                 & TRUE     \\
\bottomrule
\end{tabular}
\end{table}

\begin{table}[h]
\centering
\caption{
Training hyperparameters for adaptive interpretation training.
}
\label{tab:supp_training_hyperparameters_adaptive decoding}
\begin{tabular}{l|c}
\toprule
Parameter                & Value    \\
\midrule
Epoch                    & 1        \\
Batch Size               & 8        \\
Gradient Accumulation    & 1        \\
Optimizer                & AdamW    \\
Weight Decay             & 0.01     \\
Learning Rate            & 5e-4 \\
Learning Rate scheduler  & cosine   \\
Warmup                   & 100      \\
Clip Gradient Norm       & 1        \\
Activation Checkpointing & TRUE     \\
FSDP                     & TRUE     \\
Bfloat16                 & TRUE     \\
\bottomrule
\end{tabular}
\end{table}


\section{Additional Results}\label{sec:supp_additional_results}

\subsection{Detailed Results on MMStar}
We present the detailed accuracy results on MMStar in Table~\ref{tab:results_mmstar}, we identify that Heima outperforms Llama3.2-11B on both instance reasoning (IR) and logical reasoning (LR) tasks while using less than 10\% of the tokens, and it preserves the majority of its reasoning capabilities for mathematical problems through progressive distillation.

\begin{table}[h]
\centering
\caption{
MMStar detailed results.
CP denotes coarse perception, FP denotes fine-grained perception, IR denotes instance reasoning, LR denotes logical reasoning, S\&T denotes Science\&Technology.
Overall accuracy is a weighted metric based on sample counts.
}
\begin{tabular}{l|cccccc|c}
\toprule
Model                        & CP    & FP    & \textbf{IR}    & \textbf{LR}    & \textbf{Math}  & S \& T & Overall \\
\midrule
\begin{tabular}[c]{@{}l@{}}Llama3.2\\ -11B Vision\end{tabular} & 64.0 & 39.2 & 53.6 & 51.6 & 51.6 & 28.4 & 48.1  \\
\midrule
\begin{tabular}[c]{@{}l@{}}LLaVA-CoT\\ (reproduce)\end{tabular}         & 66.0 & 40.0 & 64.4 & 52.4 & 60.8 & 40.4 & 54.0  \\
\midrule
\midrule
\begin{tabular}[c]{@{}l@{}}Heima w/o \\ progressive\end{tabular}        & 66.0 & 43.2 & 62.4 & 45.6 & 44.8 & 36.0 & 49.7  \\
\midrule
\begin{tabular}[c]{@{}l@{}}Heima w/o \\ recover\end{tabular}            & 64.8 & 44.0 & 57.2 & 51.6 & 44.0 & 37.2 & 49.8  \\
\midrule
\textbf{Heima}                                                                   & 62.0 & 43.2 & 58.8 & 52.8 & 48.0 & 34.8 & \textbf{49.9}  \\
\bottomrule
\end{tabular}
\label{tab:results_mmstar}
\end{table}



\subsection{Detailed Evaluation of Interpreters}

We provide the detailed evaluation results for the metrics in Table~\ref{tab:supp_results_detailed_decoder_eval_metrics}.

\begin{table}[h]
\centering
\caption{
Detailed evaluation metrics for 3 interpreters trained based on LLama-3.1-8B-Instruct.
}
\label{tab:supp_results_detailed_decoder_eval_metrics}
\begin{tabular}{c|ccc}
\toprule
Stage     & Summary & Caption & Reasoning \\
\midrule
BLEU      & 15.9    & 12.8    & 11.2      \\
METEOR    & 40.1    & 35.5    & 32.7      \\
ROUGE-L   & 41.6    & 37.9    & 32.7       \\
BERTScore & 73.4    & 71.4    & 66.6      \\
\bottomrule
\end{tabular}
\end{table}

\subsection{Results with LLaVA Model Family}

We provide the results of 3 interpreters trained based on Vicuna-7B in Table~\ref{tab:supp_results_llava_decoder_eval_metrics}.


\begin{table}[h]
\centering
\caption{
Evaluation results for 3 interpreters trained based on Vicuna-7B.
}
\label{tab:supp_results_llava_decoder_eval_metrics}
\begin{tabular}{c|ccc}
\toprule
Stage     & Summary & Caption & Reasoning \\
\midrule
BLEU      & 14.5    & 13.2    & 10.6      \\
METEOR    & 39.5    & 31.5    & 31.5      \\
ROUGE-L   & 42.1    & 34.1    & 30.6       \\
BERTScore & 69.6    & 65.7    & 64.9      \\
\bottomrule
\end{tabular}
\end{table}

\subsection{Detailed Ablation Study}

We provide the detailed evaluation results of the ablation study for the different number of thinking tokens and different retention ratios in Table~\ref{tab:supp_results_detailed_ablation_num_tokens} and Table~\ref{tab:supp_results_detailed_ablation_ratio_tokens}, separately.

\begin{table*}[h]
\centering
\caption{
Detailed results for the ablation study of different number of thinking tokens.
}
\label{tab:supp_results_detailed_ablation_num_tokens}
\begin{tabular}{c|cccccc|c}
\toprule
\# Token & MMSar & MMBench & MMVet & MathVista & AI2D & Hallusion & Avg. Acc. \\
\midrule
1           & 49.9  & 72.8    & 43.3  & 43.6      & 77.5 & 60.6      & 58.0        \\
2           & 50.3  & 71.4    & 41.4  & 43.1      & 75.6 & 57.3      & 56.5        \\
4           & 49.9  & 71.0    & 42.2  & 39.3      & 75.4 & 59.3      & 56.2        \\
8           & 51.1  & 70.4    & 41.0  & 40.9      & 76.7 & 59.9      & 56.7        \\
16          & 49.5  & 72.0    & 40.9  & 40.9      & 76.2 & 61.6      & 56.9        \\
32          & 50.2  & 71.1    & 42.9  & 41.6      & 75.2 & 61.8      & 57.1       \\
\bottomrule
\end{tabular}
\end{table*}

\begin{table*}[h]
\centering
\caption{
Detailed results for the ablation study of different retention ratios.
}
\label{tab:supp_results_detailed_ablation_ratio_tokens}
\begin{tabular}{c|cccccc|c}
\toprule
Ratio & MMSar & MMBench & MMVet & MathVista & AI2D & Hallusion & Avg. Acc. \\
\midrule
0.1                              & 49.1  & 69.7    & 37.2  & 41.3      & 75.9 & 59.1      & 55.4        \\
0.2                              & 49.7  & 71.5    & 39.4  & 41.2      & 75.3 & 60.0      & 56.2        \\
0.3                              & 48.1  & 71.9    & 40.6  & 39.9      & 75.3 & 59.4      & 55.8        \\
0.4                              & 47.9  & 70.3    & 38.6  & 39.2      & 76.3 & 59.7      & 55.3        \\
0.5                              & 47.2  & 70.1    & 40.5  & 39.5      & 75.2 & 57.6      & 55.0        \\
0.6                              & 48.4  & 70.9    & 42.0  & 38.8      & 76.6 & 60.5      & 56.2        \\
0.7                              & 48.7  & 69.8    & 41.1  & 39.0      & 75.4 & 59.7      & 55.6        \\
0.8                              & 49.9  & 69.3    & 40.9  & 37.2      & 75.3 & 59.4      & 55.3        \\
0.9                              & 49.2  & 70.5    & 40.1  & 38.4      & 75.7 & 60.1      & 55.7        \\
\bottomrule
\end{tabular}
\end{table*}

\subsection{Ablation Study for Number of Decoders}\label{sec:supp_additional_ablation_num_of_decoders}
We provide additional ablation study for using one LLM as the interpreter of 3 stages with BERTScore metric in Table~\ref{tab:supp_results_ablation_number_of_decoder}. The evaluation results indicate that the single interpreter performs well on the reasoning stage but poorly on both the summary and caption stages, highlighting the necessity of employing separate interpreter for summary and caption stages.

\begin{table}[h]
\centering
\caption{
Ablation results on BERTScore metric for number of interpreters corresponding to 3 stages trained based on LLama-3.1-8B-Instruct.
}
\label{tab:supp_results_ablation_number_of_decoder}
\begin{tabular}{c|c|ccc}
\toprule
Metric                     & \# of Decoders & Summary & Caption & Reasoning \\
\midrule
\multirow{2}{*}{BLEU}      & 1              & 9.5     & 6.8     & \textbf{11.3}      \\
                           & 3              & \textbf{15.9}    & \textbf{12.8}    & 11.2      \\
                           \midrule
\multirow{2}{*}{METEOR}    & 1              & 32.7    & 25.4    & 32.5      \\
                           & 3              & \textbf{40.1}    & \textbf{35.5}    & \textbf{32.7}      \\
                           \midrule
\multirow{2}{*}{ROUGE-L}   & 1              & 36.8    & 29.7    & 31.9      \\
                           & 3              & \textbf{41.6}    & \textbf{37.9}    & \textbf{32.7}      \\
                           \midrule
\multirow{2}{*}{BERTScore} & 1              & 67.8    & 60.6    & \textbf{67.3}      \\
                           & 3              & \textbf{73.4}    & \textbf{71.4}    & 66.6     \\
\bottomrule
\end{tabular}
\end{table}

\clearpage
\section{Prompts for GPT-4o Evaluation}\label{sec:supp_gpt4o_prompts}

We provide the GPT-4o prompts in Prompt~\ref{agent:supp_gpt4o_prompts}.
In detail, we treat the evaluation as a ranking process to classify the performance of the reconstructed reasoning process into 5 ranks, from 1 to 5. 
Rank 1 represents a reconstructed reasoning process with little overlap with the ground-truth CoT and describing different themes, while Rank 5 represents a reconstruction that is well aligned with the ground truth.
We remove special tokens from both sides and input them to GPT-4o to rank the similarity at each stage. We also include the corresponding image-question pairs in the prompt as additional references for more accurate contextual support.

\definecolor{mycolor}{RGB}{55, 71, 108}
\definecolor{lightgraybox}{RGB}{245,245,245}

\tcbset{
  agentboxstyle/.style={
    enhanced,
    breakable,              
    arc=2mm, 
    colback=lightgraybox,
    colframe=black,
    boxrule=1pt,
    width=\textwidth,
    before skip=10pt,
    after skip=10pt,
    coltitle=white,
    fonttitle=\bfseries,
    colbacktitle=mycolor
  }
}
\newtcolorbox{agentbox}[1]{agentboxstyle, title={#1}}

\begin{agentbox}{CoT Reconstruction Evaluation Agent (GPT-4o)}

\textbf{Input:} Image $\mathbf{I}$, question $\mathbf{Q}$, reconstructed CoT $\hat{\mathbf{CoT}}$, ground-truth CoT $\mathbf{CoT}$, and CoT stage type $\mathbf{T} \in \{\text{caption}, \text{summary}, \text{reasoning}\}$.

\textbf{Output:} An integer similarity rank in $\{1,2,3,4,5\}$ measuring the alignment between $\hat{\mathbf{CoT}}$ and $\mathbf{CoT}$.

\noindent\rule{\linewidth}{0.4pt}




\vspace{0.3em}
\textbf{Agent Role}

You are an evaluation agent responsible for assessing whether a generated analysis aligns with the ground truth, given the image–question pair.  
The analysis corresponds to one of the following stages:
\begin{itemize}[leftmargin=*]
    \item \textbf{Caption}: Description of the image content.
    \item \textbf{Summary}: Restatement or paraphrase of the question.
    \item \textbf{Reasoning}: Logical explanation linking the image and question to the answer.
\end{itemize}

\vspace{0.3em}
\textbf{Evaluation Task}

You are given:
\begin{itemize}[leftmargin=*]
    \item Generated stage $\mathbf{T}$: $\hat{\mathbf{CoT}}$
    \item Ground truth: $\mathbf{CoT}$
    \item Image $\mathbf{I}$ and question $\mathbf{Q}$
\end{itemize}

Evaluate how closely the generated $\mathbf{T}$ matches the ground truth according to the image and question.

\vspace{0.3em}
\textbf{Similarity Scale}

\begin{itemize}[leftmargin=*]
    \item \textbf{1 — Completely unrelated}: No thematic or factual overlap.
    \item \textbf{2 — Minimally related}: Only weak or tangential overlap.
    \item \textbf{3 — Partially related}: Core theme overlaps but with clear factual errors or missing key details.
    \item \textbf{4 — Closely related}: Main content aligns with minor discrepancies.
    \item \textbf{5 — Nearly identical}: Almost complete alignment with negligible differences.
\end{itemize}

\vspace{0.3em}
\textbf{Response Format}

Return a JSON object:
\(
\{\;
"\mathbf{T}": \text{Rank},\;
"reason": \text{explanation}
\;\}
\)
where \texttt{Rank} is the integer similarity score and \texttt{reason} briefly justifies the decision.

\end{agentbox}
\captionof{figure}{Instruction prompts for the Reconstruction Evaluation Agent using GPT-4o.}
\label{agent:supp_gpt4o_prompts}

\end{document}